\documentclass[sigconf,review=false,anonymous=false,authordraft=false]{acmart}

\copyrightyear{2024}
\acmYear{2024}
\setcopyright{rightsretained}

\acmConference[SIGIR '24]{Proceedings of the 47th International ACM SIGIR Conference on Research and Development in Information Retrieval}{July 14--18, 2024}{Washington, DC, USA.}
\acmBooktitle{Proceedings of the 47th Int'l ACM SIGIR Conference on Research and Development in Information Retrieval (SIGIR '24), July 14--18, 2024, Washington, DC, USA}

\makeatother

\usepackage{enumitem}
\newlist{inlinelist}{enumerate*}{1}
\setlist*[inlinelist,1]{%
  label=(\roman*),
}
\usepackage{subfigure}

\newcommand{\ltr}{ROPG\xspace}
\newcommand{\ltrrl}{ROPG-RL\xspace}
\newcommand{\ltrkd}{ROPG-KD\xspace}
\newcommand{\lts}{RSPG\xspace}

\title{Optimization Methods for Personalizing Large Language Models through Retrieval Augmentation}

\author{Alireza Salemi}
\affiliation{\institution{University of Massachusetts Amherst}
\country{United States}}
\email{asalemi@cs.umass.edu}

\author{Surya Kallumadi}
\affiliation{\institution{Lowe’s Companies, Inc.}
\country{United States}}
\email{surya@ksu.edu}

\author{Hamed Zamani}
\affiliation{\institution{University of Massachusetts Amherst}
\country{United States}}
\email{zamani@cs.umass.edu}

\usepackage{multirow}
\usepackage{adjustbox}
\usepackage{amsmath}
\usepackage{xspace}

\begin{document}


\begin{abstract}
This paper studies retrieval-augmented approaches for personalizing large language models (LLMs), which potentially have a substantial impact on various applications and domains. We propose the first attempt to optimize the retrieval models that deliver a limited number of personal documents to large language models for the purpose of personalized generation. We develop two optimization algorithms that solicit feedback from the downstream personalized generation tasks for retrieval optimization--one based on reinforcement learning whose reward function is defined using any arbitrary metric for personalized generation and another based on knowledge distillation from the downstream LLM to the retrieval model. This paper also introduces a pre- and post-generation retriever selection model that decides what retriever to choose for each LLM input. Extensive experiments on diverse tasks from the language model personalization (LaMP) benchmark reveal statistically significant improvements in six out of seven datasets.
\end{abstract}





\maketitle

\section{Introduction}
\label{sec:intro}

Personalization has been extensively explored by information retrieval (IR), recommender systems, and human-computer interaction communities, particularly in the context of information access \cite{10.1145/2702123.2702503, 10.1145/1462198.1462203, naumov2019deep}. Even though the recent advancements in large language models (LLMs) have revolutionized various applications, the existing commercial and open-source LLMs exhibit a significant limitation by failing to tailor their generated outputs according to the backgrounds and historical preferences of their users. As LLM-powered conversational agents become more prevalent, the need for \emph{LLM personalization} becomes increasingly apparent \cite{lamp}. LLM personalization has diverse applications, from customizing educational content and curating news feeds to improving e-commerce suggestions and delivering personalized healthcare information.

\begin{figure}[t]
    \centering
    \includegraphics[width=.8\linewidth]{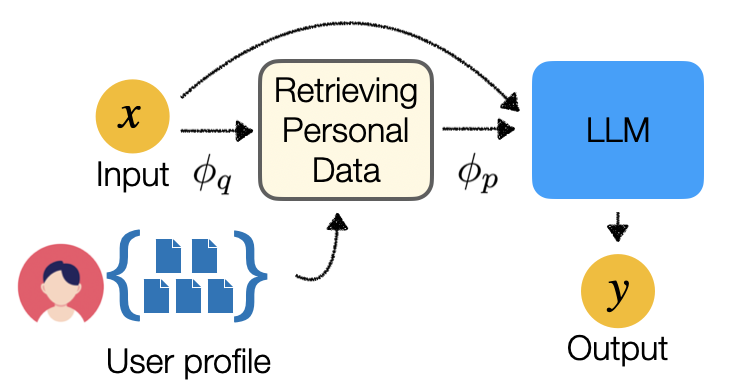}
    \caption{An overview of retrieval augmentation approaches for LLM personalization. First, the query function $\phi_{q}$ produces a query from the input $x$. Relevant personal information is then retrieved and fed to the personalized prompt generation function $\phi_{p}$ for LLM consumption.}
    \label{fig:rag4personalization}
    \vspace{-0.6cm}
\end{figure}

Various approaches can be envisioned to personalize LLMs: (1) fine-tuning LLM parameters, either entirely or partially, for individual users, (2) integrating latent user representations with LLMs, and (3) enriching LLM prompts with user-specific content and/or context. The first two approaches involve adjusting LLM architecture and parameters, which is costly or even impractical in terms of storage, computation cost, and/or time. Besides, they cannot perform well for cold-start users. As an instantiation of retrieval-enhanced methods \cite{reml}, the third approach, on the other hand, is applicable to any off-the-shelf LLM. To efficiently and effectively utilize the potentially extensive personal data for each active user, it is essential to implement a retrieval mechanism. As presented in \figurename~\ref{fig:rag4personalization}, this mechanism selects personal information that best enhances the LLM for the purpose of personalized text generation.

This paper focuses on optimization of personal information retrieval for the purpose of personalizing LLMs. For this purpose, standard learning-to-rank optimization methods are not applicable, since they typically require query-document-relevance triplets for training and it is not clear what documents is ``relevant'' for the downstream personalized text generation tasks. We study two retrieval optimization solutions for personalizing LLMs. First, we develop a reinforcement learning approach, where the training process involves sampling documents from the user's profile with respect to the probabilities generated based on retrieval scores. The sampled documents are then fed to the LLM and its downstream performance for producing personalized output texts (using any arbitrary metric) is used to compute a reward function for optimizing the retrieval model. 
Second, we optimize the retrieval model by distilling knowledge from the LLM. We minimize the divergence between the retrieval score distribution and a target distribution computed using the downstream performance of LLM on producing personalized output texts using each individual retrieved document. To gauge the efficacy of these two optimization methods, we apply them to train dense retrieval models for LLM personalization.


Moreover, we observe that LLM personalization has multiple dimensions and existing retrieval models do not address all of them. For instance, retrieving \emph{recent} user interactions may be needed for effective personalized text generation for an input, while a keyword-matching, a semantic matching, or a model that is aware of user's writing style may be optimal for another input. According to this observation, we hypothesize that a retrieval selection model that selects a model from a diverse pool of retrievers can impact LLM personalization. Following this hypothesis, we further develop \emph{a pre- and a post-generation model mode for retrieval selection} that decides what retrieval model should be chosen for each given input. In our study, these models could choose between (1) no retrieval (i.e., no personalization), (2) recency-based retrieval, (3) term matching (i.e., BM25 \cite{bm25}), (4) zero-shot semantic matching (Contriever \cite{contriever}), and (5, 6) the two dense retrieval models developed in this paper that are specifically trained for LLM personalization.  
To optimize the retrieval selection models, we align the probability distribution obtained from the LLM downstream performance and the scores generated by the selection model for various retrieval models.

We evaluate our models using the LaMP benchmark \cite{lamp}, consisting of seven diverse personalization tasks, including three personalized text classification (binary, categorical, and ordinal) and four personalized text generation tasks. The methods proposed in this paper advance the state-of-the-art performance on six out of seven tasks in LaMP with statistically significant improvements. Our best-performing method exhibits an average of 5.5\% state-of-the-art improvements across all LaMP datasets. Comparing to the non-personalized LLM, our best approach demonstrate 1.0\%-33.8\% improvements across all tasks, with an average improvement of 15.3\%. To facilitate the research on this domain, we share our codes and trained model parameters to support future research, promoting transparency and reproducibility. \footnote{\url{https://github.com/LaMP-Benchmark/LaMP}}



\section{Related Work}

\subsubsection*{\textbf{Personalized Text Generation}}

Personalization has been a focal point of research in various domains, particularly within search and recommendation systems \cite{Bennett:2012, Dumais:2016, citeulike:2187446, PERSON:2018, uia}. This exploration spans diverse contexts, encompassing areas such as query auto-completion \cite{personalized-query} and collaborative personalized search \cite{10.1145/1462198.1462203}. Within the NLP community, personalization has been a subject of exploration in various applications, including but not limited to dialogue agents \cite{dialog2, dialog3, mazare2018training, zhong2022less, Qian2021pchatbot, vincent2023personalised}, review\cite{lituzhilin2019towards} and recipe generation \cite{majumder2019generating}, translation\cite{wuebker2018compact}, headline generation \cite{ao2021pens}, and classification tasks \cite{flek2020returning, dudy2021refocusing}, such as personalized sentiment analysis \cite{mireshghallah2022useridentifier}.

With the emergence of LLMs and their application across various NLP tasks, \citet{lamp} proposed a retrieval-augmented approach for personalizing LLMs. They also introduced LaMP, a benchmark designed to assess the performance of personalized NLP models across diverse classification and short text generation tasks. The work by \citet{Li2023TeachLT} addresses a similar issue, focusing on personalized long text generation. Furthermore, \citet{Mysore2023PEARLPL} assesses the capabilities of LLMs in the role of writing assistants. Various approaches have been explored for personalizing LLMs, encompassing techniques such as summarizing user profile \cite{Richardson2023IntegratingSA}, aligning language models with personalized human feedback \cite{Jang2023PersonalizedSP}, automatic prompt generation tailored to individual users \cite{Li2023AutomaticPR}, and incorporating long and short-term memory-based personalization strategies \cite{Zhang2023MemoryAugmentedLP}. In this study, we adhere to the methodology outlined by \citet{lamp} and conduct experiments using the LaMP benchmark, focusing on training a component for retrieving personal information from user profile.

\subsubsection*{\textbf{Retrieval Optimization in Retrieval-Augmented Generation}}

The optimization of retrieval models within the RAG (Retrieval-Augmented Generation) pipelines has emerged as a focal point in recent research, particularly in the context of question answering. \citet{yang2020retriever} focuses on distilling knowledge from the LM to the retriever by minimizing the KL-divergence between the LM's performance for each document in the retrieved set and the assigned score by the retriever to that document in the set. Additionally, \citet{izacard2021distilling} employs the attention weights of the LM to determine the importance of each document. This information is then utilized to distill knowledge from the LM to the retriever, aligning with the objectives set forth by \citet{yang2020retriever}. The approach presented by \citet{wang2023retrieve} involves the use of reinforcement learning, where the reward function is derived from the performance of the LM. In this work, we adopt methods similar to that of \citet{wang2023retrieve} and \citet{yang2020retriever}, given the absence of relevance data in the LaMP benchmark. Notably, our work stands out as the pioneering effort in leveraging feedback from LLMs to train personalized retrievers for personalizing LLMs. Furthermore, in all the previously mentioned approaches, the language model is trained after/with the retrieval model. In contrast, our approach assumes the language model is frozen, and our focus is solely on optimizing the retrieval model. 

\subsubsection*{\textbf{Information Access with Multiple Retrieval Models}}

Combining rank lists generated by different retrievers has been extensively explored in the literature \cite{NURAY2006595, LOSADA201856, 10.1145/1277741.1277843, rrf}. However, the process of rank fusion presents challenges, especially when dealing with discrepancies in scoring scales among retrieval systems or the absence of overlapping documents in the ranked lists \cite{LOSADA201856, 6873347}. Alternatively, methods for selecting specific retriever from a retriever pool for different datasets has been explored \cite{khramtsova2023selecting}. Furthermore, \citet{10.1145/3459637.3482159} investigates the optimal use of dense and sparse retrievers for each query, considering efficiency trade-offs. In our study, we concentrate on the performance-oriented selection of query-specific retrievers from a retriever pool.

\section{Notations and Task Formulation}
\label{sec:definition}
Generative language models often take an input $x$ and generate the most probable sequence tokens $y$. This paper focuses on the task of personalized generation with the goal of generating outputs that are tailored for the preferences and characteristics of the language model user. Let $T = \{(u_1, x_1, y_1), (u_2, x_2, y_2), \cdots, (u_N, x_N, y_N)\}$ be a set of $N$ training instances, each consisting of a user $u$, an input text $x$ submitted by the user $u$, and the ground truth personalized output $y$. For each user $u$, a user profile $P_u$ exists that can be employed for developing personalized generation models. A user profile $P_u$ is a set of personal documents associated with the user $u$. 

As discussed in Section~\ref{sec:intro}, this paper focuses on retrieval augmented solutions for personalization, depicted in \figurename~\ref{fig:rag4personalization}. In such solutions, we first retrieve a set of personal documents from the user profile $P_u$. This is achieved through $L = \mathcal{R}(\phi_q(x); P_u)$ where $\phi_q$ is a query generation function that produces a search query string given the LLM input $x$ and $\mathcal{R}$ is a retrieval model that retrieves personal documents from $P_u$ given a query produced by $\phi_q$. Hence, $\mathcal{R}$ returns a list of personal documents $L$. A prompt generation function $\phi_p$ is then applied to the LLM input and the retrieved result list as follows: $\phi_p(x, L)$. The constructed personalized prompt is then fed into a LLM $M$. The goal is to minimize the error between the generated output and the ground truth personalized output $y$. We assume that the LLM $M$ is given and we do not aim at updating the LLM parameters for personalized generation. The rational behind this decision is that (1) fine-tuning $M$ is often very expensive, and more importantly (2) the LLM $M$ can memorize personal information if it is fine-tuned on data retrieved from the user's personal data $P_u$. Such memorization can put the user's privacy at risk. That said, this paper focuses on minimizing the personalized text generation error by solely updating the retrieval results $L$. 

Section~\ref{sec:personalized-retriever} studies methods for optimizing the retrieval model $\mathcal{R}_\theta$ parameterized by $\theta$ for updating the result list $L$. Section~\ref{sec:retriever-model-selection} extends Section~\ref{sec:personalized-retriever} by exploring optimization solutions for retrieval model selection from a set of pre-defined retrieval models for updating the result list $L$.

\section{Learning to Retrieve for Personalizing LLMs}
\label{sec:personalized-retriever}
Learning-to-rank (LTR) methods are often employed to train ranking models for search and recommendation \cite{ltr}. For personalized LTR, the user's profile and long-term history are often utilized. Such personalized implicit feedback signals are often document-level and directly provided by the user, such as ratings, clicks, views, dwell time, and/or purchases \cite{uia,Bennett:2012}. 
In the context of retrieval-augmented personalized text generation, user feedback manifests in the form of text written or edited by the user (i.e., the label $y_i$ for each input $x_i$), taking into account the user preferences and interests. Therefore, accessing user feedback on a per-document basis within the user profile for training retrieval models is not feasible; neither is collecting document-level feedback through annotation, e.g., crowdsourcing. The main reason is that we do not know what documents serve the LLM best for generating personalized outputs for each input text. Therefore, learning to rank documents for LLM personalization is fundamentally different from developing personalized search or recommendation engines.




Considering these, we propose optimization methods that leverage feedback from the LLM itself, obtained by evaluating the impact of retrieved documents on the LLM performance for generating personalized outputs. Our first method uses the LLM performance (in terms of any arbitrary metric) to form a reward function and employs reinforcement learning \cite{reinforce} for optimization. Our second approach is based on knowledge distillation from the LLM to the retriever based on the LLM's performance in terms of personalized text generation. They are described in the subsequent subsections.

\subsection{Retrieval Optimization for Personalized Generation using Reinforcement Learning}
\label{sec:training-rl}

\begin{figure*}
    \centering
    \includegraphics[width=0.95\textwidth]{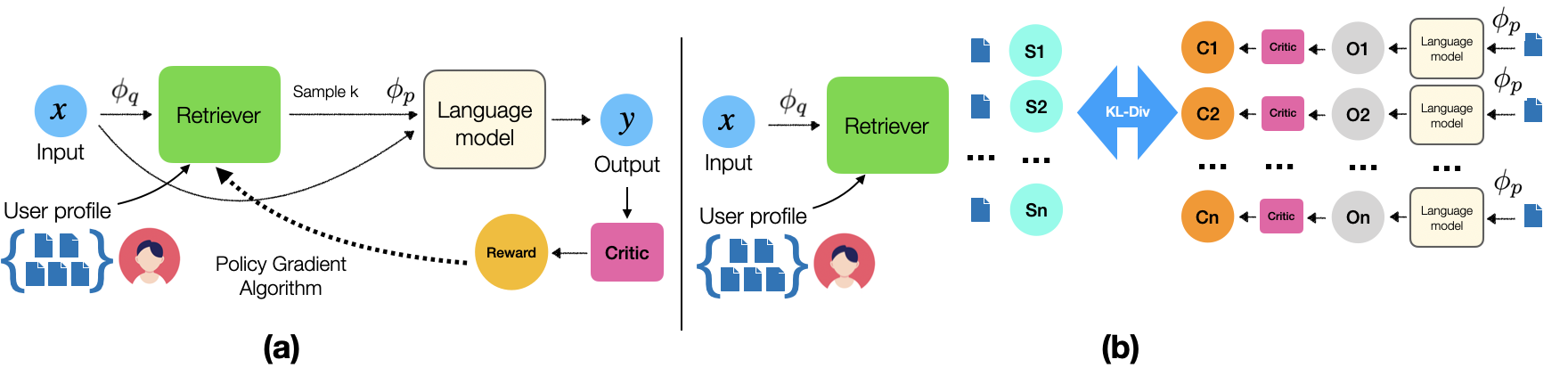}
    \vspace{-0.4cm}
    \caption{Overview of training dense retrievers for personalizing LLMs using LLMs feedback with policy gradient optimization (a) and knowledge distillation (b). $\phi_q$ represents the query generation function, $\phi_p$ is the prompt generation function, and "Critic" denotes the evaluation metric employed for the personalized task.}
    \label{fig:train-rl-kl}
\end{figure*}

This section introduces \ltrrl--a reinforcement learning approach that encourages the retrieval model to produce rankings that lead to more accurate personalized generation. We utilize the vanilla policy gradient optimization algorithm, drawing upon rewards supplied by the downstream LLM, as depicted in Figure \ref{fig:train-rl-kl} (a). 
In this approach, we establish a parameterized policy (here, the retrieval model) that assigns a probability to each action (specifically, selecting personal documents to be fed to the LLM). Subsequently, we formulate a reward function, well aligned with the goal of personalized text generation, that the parameterized policy aims to maximize. This approach enables us to effectively refine and improve the performance of the retrieval model. The specifics of our methodology will be discussed in the subsequent paragraphs.

\noindent
\subsubsection*{\textbf{Parameterized Policy (${\pi_\theta}$)}} 
In this context, defining the policy function necessitates a clear delineation of the states and actions applicable within the scope of this study. In this formulation, \emph{the policy is parameterized through the retrieval model.} Essentially, given a query, the retrieval model undergoes training to assign a higher probability to the documents from the user profile that are deemed more valuable for personalizing the LLM.
Here, an action is considered as selecting a document given a query.\footnote{We also explored scenarios where multiple documents are sampled without replacement from the policy network, however, we observed no or little improvement compared to a much simpler and more efficient approach where only one document is sampled for updating the policy parameters.} The state, on the other hand, corresponds to the given query itself. Meaning that the goal is to update the policy such that it produces more effective results for personalization. In this study, we restrict our focus to trajectories comprising a single state. This implies that the model initiates from the initial state, executes a singular action, and concludes the trajectory. The probability of each action is computed using the following formula:

\begin{equation}
    \label{eq:rl-prob-action}
    \pi_{\theta}(d | x) = \frac{\exp{(\mathcal{R}_\theta(\phi_q(x), d))}}{\sum_{d' \in {P_u}} \exp{(\mathcal{R}_\theta(\phi_q(x), d'))}} \quad : \quad \forall d \in P_u
\end{equation}
where $\phi_q$ is the query generation function as explained in Section \ref{sec:definition}, $\mathcal{R}_{\theta}$ denotes the retrieval model parameterized by $\theta$, and $P_u$ is a set of documents containing user's personal data (see Section~\ref{sec:definition}). The probability assigned by the policy model to any document that is not in the user profile would be zero. Note that the user profile can grow over time that leads to inefficient calculation of policy function. To address this efficiency issues, we can approximate the policy function, either using hierarchical softmax, similar to \cite{Morin:2005,Mikolov:2013,Zamani:2017}, or by marginalization through top $l$ approximation. Without loss of generality we choose the second approach and compute the policy function based on $P_u^l$, a set of $l$ documents from $P_u$ that achieve highest retrieval scores according to our initial retrieval weights. In our experiments, we set $l=16$. Of course, leveraging the complete user profile during training could potentially yield superior performance, albeit at the cost of increased training time.




\noindent
\subsubsection*{\textbf{Reward Function (${R}$)}} 
Defining trajectories involving the selection of multiple documents from the user profile, with rewards for each, is computationally intensive. This is due to the need to compute rewards using the LLM for every sampled set of documents from the user profile. On the contrary, if we limit trajectories to selecting only one document from the user profile, we can pre-compute the rewards for each document in the profile. Therefore, we only consider trajectories with one action. Moreover, to expedite the learning process and minimize variance, we subtract a previously calculated evaluation score from the effectiveness of the current sampled document. The reward function is then defined as:
\begin{align}
    \label{eq:reward-function}
    \textsc{Reward}&(d; x, y) =  \\
    &\textsc{Eval}\left(y, M(\phi_p(x, [d]))\right) - \textsc{Eval}\left(y, M(\phi_p(x, [d_b]))\right) \nonumber
\end{align}
where $d$ is a sampled document from the user profile $P_u$ using the parameterized policy $\pi_\theta$ and $d_b$ is the document selected by the policy with initial weighting (i.e., the retrieval model prior to fine-tuning with RL). $M$ is the LLM that generates personalized text based on the given personalized prompt $\phi_p$. $\textsc{Eval}$ denotes an arbitrary metric for evaluating personalized text generation.

\noindent
\subsubsection*{\textbf{Training Objective (${J}$)}}
The objective in \ltrrl is to maximize the expected reward obtained by the parameterized policy of the retriever ($\pi_\theta$). To achieve this objective, we employ a gradient ascent algorithm with the update rule of $\theta_{k+1} = \theta_{k} + \alpha \nabla_{\theta}J(\pi_\theta)$. For each mini-batch $B \subset T$, the objective function is presented below:
\begin{equation} 
\arg \max_\theta \frac{1}{|B|} \sum_{(u, x, y) \in B} \mathop{\mathbb{E}}_{d \sim \pi_\theta}\left[\textsc{Reward}(d; x, y) \log \pi_{\theta}(d|x)\right] 
\end{equation}



\begin{figure*}
    \centering
    \includegraphics[width=0.99\textwidth]{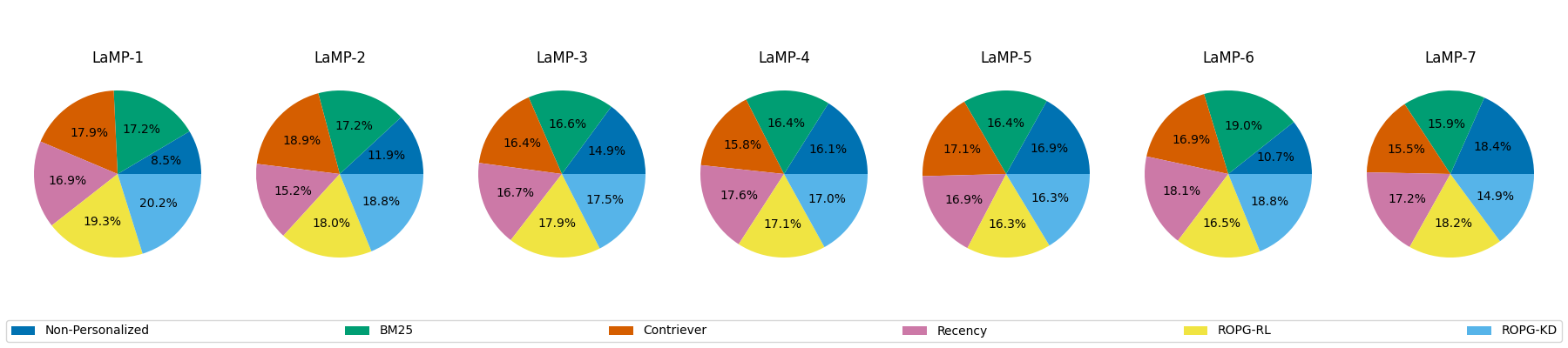}
    \caption{Relative winning rate for each selected retrieval model. When multiple retrieval models get the highest score, we consider all of them with the highest score as the winner.}
    \label{fig:retriever-win-rate}
\end{figure*}

\subsection{Retrieval Optimization for Personalized Generation using Knowledge Distillation}

An alternative approach for training a retrieval model with feedback from the LLM involves knowledge distillation from the LLM to the retrieval model. Contrary to \ltrrl, this approach, called \ltrkd, considers the relative usefulness of different items in the user profile for the LLM on performing the downstream personalized task. Indeed, this approach endeavors to allocate a higher probability to items that are more useful than others for the LLM. 

Conversely, \ltrrl only considers the impact of each (or possibly a subset of documents grouped together in trajectories with more than one action) on the final score that the LLM achieves. Hence, it seeks to reward the model for actions that yield positive outcomes and penalize for actions that result in negative consequences. This implies that where there are no favorable actions, the model would face punishment; however, this is not the case with knowledge distillation. On the other hand, RL optimization processes tend to be less stable and are more susceptible to overfitting. 

Considering all the aforementioned aspects, we propose an alternative approach based on knowledge distillation. In the context of knowledge distillation, the primary objective is to encourage the retriever model to assign higher similarity scores to the documents from the user profile that are more useful for the language model in fulfilling its task. Figure~\ref{fig:train-rl-kl} (b) illustrates the pipeline for this approach. To accomplish this objective, we employ Equation \eqref{eq:rl-prob-action} to allocate a probability to individual elements within the user profile. Subsequently, we use the following function to produce the target probability distribution:
\begin{equation*}
    p_t(d | x) = \frac{\exp{(\textsc{Eval}\left(y, M(\phi_p(x, [d]))\right))}}{\sum_{d' \in {P_u}} \exp{(\textsc{Eval}\left(y, M(\phi_p(x, [d']))\right))}} \quad : \quad \forall d \in P_u
\end{equation*}
where $\textsc{Eval}$ is an arbitrary metric for evaluating personalized text generation models and $M$ denotes the LLM being used. 
Similar to \ltrrl, for efficiency purposes, we approximate the distribution $p_t$ by only focusing on the top $l$ retrieved document w.r.t. the initial retrieval parameters.  Inspired by previous work on knowledge distillation in IR \cite{yang2020retriever}, for each mini-batch $B \subset T$, we minimize the following loss function based on KL-divergence:
\begin{equation}
    \arg \min_\theta \frac{1}{|B|} \sum_{(u, x, y) \in B} \sum_{d \in P^l_u} p_t(d | x) \log\frac{\pi_{\theta}(d|x)}{p_t(d | x)}
\end{equation}


\subsection{Retrieval Model Architecture}
The proposed optimization solutions can be applied to any neural ranking model. Without loss of generality, we use them to train dense retrieval models. We adopt Contriever \cite{contriever}, a pre-trained bi-encoder model for dense retrieval. Contriever encodes the query and document text using an encoder with shared parameters and applies dot product to compute the relevance score. After training, an exact or approximate nearest neighbor (kNN) algorithm is used to index the learned document representations for each user profile. We use exact kNN in our experiments. During inference, each document is scored independently and the scored documents are sorted in descending order with respect to their score.




\section{Retrieval Model Selection for Personalizing LLMs}
\label{sec:retriever-model-selection}

We hypothesize that there are multiple aspects to LLM personalization and each existing retrieval model does not address all of them. For instance, retrieving recent user interactions may lead to the highest personalized text generation performance for an input, while a keyword-matching model or a semantic matching model may be optimal for another input. To validate this, we utilize the LaMP benchmark--a recent benchmark consisting of seven diverse tasks for training and evaluating personalized LLMs \cite{lamp}. Statistics of these datasets are presented in \tablename~\ref{tab:task-stats}. More information on the LaMP benchmark is provided in Section~\ref{sec:exp:setup}. We evaluate personalization of an 11B parameter FlanT5-XXL \cite{flant5} using the following retrieval augmentation approaches: (1) no personalization (i.e., no retrieval augmentation), (2) term matching retrieval using BM25 \cite{bm25}, (3) zero-shot semantic matching model using Contriever \cite{contriever}, (4) ranking documents in the user profile based on recency, and (5 \& 6) the proposed retrieval models for LLM personalization--\ltrrl and \ltrkd. \figurename~\ref{fig:retriever-win-rate} illustrates the winning rate achieved by each of these models when retrieving from the user profile. To compute the winning rate, we first evaluate the recommended evaluation metric for each of the datasets by the LaMP benchmark (i.e., accuracy for categorical classification, MAE for ordinal classification, and Rouge-1 for text-generation tasks). For each retrieval model, we then count the number of inputs for which it achieves the highest personalized text generation performance among the mentioned six models. If two or more retrieval models achieve the same and the highest performance, they all get rewarded. The count is then normalized to estimate the winning rate.

As evident in \figurename~\ref{fig:retriever-win-rate}, there is no consistent winner across all tasks in the LaMP benchmark. For instance, between the best text generation performance for 8.5\% to 18.4\% of the inputs across datasets can be achieved when no personalization is conducted. Recency-based ranking can lead to the best performance for 15.2\% to 18.1\% of the inputs depending on the dataset. Even the proposed \ltrrl and \ltrkd models provide the highest performance for 14.9\% to 20.2\% of the inputs across different datasets in LaMP. 

Given these observations, we hypothesize that selecting what ranking function to use for each input, or even when to apply personalization, can improve end-to-end personalized text generation performance. According to this hypothesis and inspired by the query performance prediction literature \cite{Carmel:2010,Zamani:2018,Shtok:2012,Roitman:2017}, we introduce two retriever selection models. In the first approach, referred to as \lts-Pre, we retrieve items from the user profile using each retriever in the retriever pool to construct personalized prompts. These constructed prompts are then fed into \lts-Pre for selection and consumption by the LLM. In the second approach, termed \lts-Post, the personalized prompts produced by all retrieval models are also presented to the LLM, and the resulting outputs, along with the original prompts, are fed into \lts-Post for retrieval model selection. These models are presented below.

\noindent \subsubsection*{\textbf{Optimizing Retriever Selection Models}}

\begin{figure*}
    \centering
    \includegraphics[width=.9\textwidth]{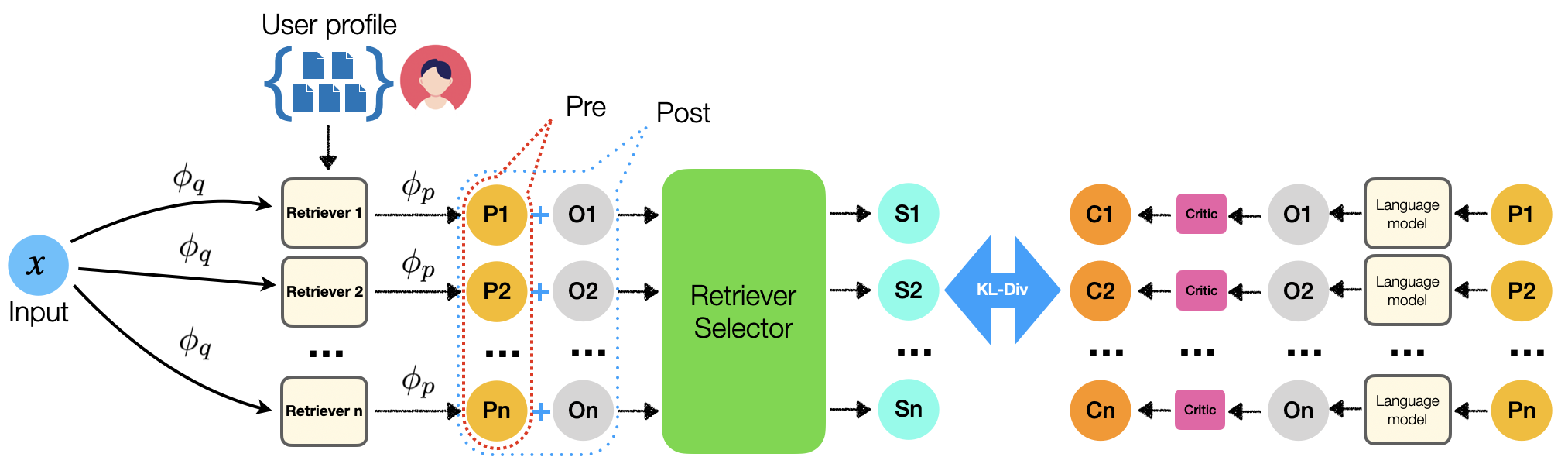}
    \caption{Pipeline for training the \lts models. We minimize the KL-divergence between the scores generated for each prompt using \lts and the performance of the prompt used for the evaluation of the LLM in a personalized task.}
    \label{fig:train-retriever-selector}
\end{figure*}

\begin{table*}
    \centering
    \caption{Statistics of the datasets within the LaMP benchmark \cite{lamp} with time-based data separation.}
    \vspace{-0.3cm}
    \begin{adjustbox}{max width=\textwidth}    
        \begin{tabular}{l|ccc|cc|c|c}
        
            \textbf{Task}  & \textbf{\#train} & \textbf{\#dev} & \textbf{\#test} & \textbf{Input Length} & \textbf{Output Length} & \textbf{\#Profile Size} & \textbf{\#classes}\\
            \hline
            {LaMP-1: Personalized Citation Identification}  & 6542 & 1500 & 1500 & 51.43 $\pm$ 5.70 & - & 84.15 $\pm$ 47.54 & 2 \\
            \hline
            {LaMP-2: Personalized Movie Tagging}  & 5073 & 1410 & 1557 & 92.39 $\pm$ 21.95 & - & 86.76 $\pm$ 189.52 & 15 \\
            \hline
            {LaMP-3: Personalized Product Rating}  & 20000 & 2500 & 2500 & 128.18 $\pm$ 146.25 & - & 185.40 $\pm$ 129.30 & 5 \\
            \hline
            {LaMP-4: Personalized News Headline Generation} & 12500 & 1500 & 1800 & 29.97 $\pm$ 12.09 & 10.07 $\pm$ 3.10 & 204.59 $\pm$ 250.75 & - \\
            \hline
            {LaMP-5: Personalized Scholarly Title Generation} & 14682 & 1500 & 1500 & 162.34 $\pm$ 65.63 & 9.71 $\pm$ 3.21 & 87.88 $\pm$ 53.63 & - \\
            \hline
            {LaMP-6: Personalized Email Subject Generation}  & 4821 & 1250 & 1250 & 454.87 $\pm$ 889.41 & 7.37 $\pm$ 2.78 & 55.67 $\pm$ 36.32 & - \\
            \hline
            {LaMP-7: Personalized Tweet Paraphrasing} & 13437 & 1498 & 1500 & 29.72 $\pm$ 7.01 & 16.96 $\pm$ 5.67 & 15.71 $\pm$ 14.86 & - \\\hline
        \end{tabular}
    \end{adjustbox}
    \label{tab:task-stats}
\end{table*}

The training pipeline for retrieval selection is illustrated in Figure \ref{fig:train-retriever-selector}. Let $\mathbf{R}$ be a set of retrieval models in the pipeline and $\mathcal{S}_\omega$ be a retrieval selection model parameterized by $\omega$ that produces a selection score for each retrieval model in $\mathbf{R}$. We use a knowledge distillation loss from the downstream LLM performance to train the retrieval selection model. For this purpose, the target selection probability distribution over retrieval models in $\mathbf{R}$ for an input $(u, x, y)$ is computed as:
\begin{equation}
    P_{TS}(\mathbf{R}_i | x; u, y) = \frac{\exp{(\textsc{Eval}\left(y, M(\phi_p(x, \mathbf{R}_i(\phi_q(x); P_u)))\right))}}{\sum_{j=1}^{|\mathbf{R}|} \exp{(\textsc{Eval}\left(y, M(\phi_p(x, \mathbf{R}_j(\phi_q(x); P_u)))\right))}}
\end{equation}
where $\textsc{Eval}$ is an arbitrary evaluation metric for personalized text generation, and $M$ is the LLM used for text generation. The retrieval selection model probability for the retriever $\mathbf{R}_i \in \mathbf{R}$ is calculated as:
\begin{equation}
    P_{S_\omega} (\mathbf{R}_i | x) = \frac{\exp(\mathcal{S}_\omega(\mathbf{R}_i, x))}{\sum_{j=1}^{|\mathbf{R}|}\exp(\mathcal{S}_\omega(\mathbf{R}_j, x))}
\end{equation}

For a mini-batch $B \subset T$, we use KL-divergence loss as follows to train the retrieval selection models:
\begin{align}
    \frac{1}{|B|} \sum_{(u, x, y) \in B}
    \sum_{i=1}^{|\mathbf{R}|} P_{TS}(\mathbf{R}_i | x; u, y) \log \frac{P_{S_\omega} (\mathbf{R}_i | x)}{P_{TS}(\mathbf{R}_i | x; u, y)}
\end{align}

\noindent \subsubsection*{\textbf{Selection Model Architecture}}
To select a retrieval model for the given input, we initiate the process by employing all retrieval models (see \figurename~\ref{fig:retriever-win-rate} for the list of retrieval models in our experiments) to retrieve relevant documents. As mentioned earlier, we envision two categories of retrieval selection models: pre-generation and post-generation models. We use an encoder-only model for estimating a selection score for each retrieval model. For the pre-generation scenario, the input to the encoder is the LLM prompt constructed by each retrieval model: $\phi_p(x, \mathbf{R}_i(\phi_q(x); P_u))$. For the post-generation scenario, this prompt is concatenated with the LLM's output text given this prompt. 
The encoder's final layer representation is then fed into a linear projection layer for producing a scalar value as the selection score. Given the long length of prompts in retrieval selection, we use Longformer \cite{beltagy2020longformer} as the encoder in our retrieval selection model.\footnote{\url{https://huggingface.co/allenai/longformer-base-4096}} Once all retrieval models are scored using this approach, we choose the one with the highest selection score and feed the corresponding prompt to the LLM.

\begin{table*}[!t]
    \centering
    \caption{Templates used to create prompt to augment the input for LLM with the retrieved items from the user profile (i.e., $\phi_p$). Function \textcolor{blue}{concat} concatenates the strings in its first argument by placing the second argument between them. Function \textcolor{blue}{add\_to\_paper\_title} adds the string in its first argument to the paper's title in the LaMP-1 task. Function \textcolor{blue}{PPEP} creates the prompt for each retrieved item from the profile. \textcolor{red}{[INPUT]} is the task's input ($x$).}
    \vspace{-0.2cm}
    \begin{adjustbox}{max width=\textwidth}
    \begin{tabular}{l|p{5cm}|p{10cm}}
        \textbf{Task} & \textbf{Per Profile Entry Prompt (PPEP)} & \textbf{Aggregated Input Prompt (AIP)} \\
        \hline
            LaMP-1: Citation Ident. & "$P_i$[title]" & \textcolor{blue}{add\_to\_paper\_title}(\textcolor{blue}{concat}([\textcolor{blue}{PPEP}($P_1$), ..., \textcolor{blue}{PPEP}($P_n$)], \textcolor{gray}{", and "}), \textcolor{red}{[INPUT]}) \\
            LaMP-2: : Movie Tag. & the tag for the movie: "$P_i$[description]" is "$P_i$[tag]" & \textcolor{blue}{concat}([\textcolor{blue}{PPEP}($P_1$), ..., \textcolor{blue}{PPEP}($P_n$)], \textcolor{gray}{", and "}). \textcolor{red}{[INPUT]} \\
            LaMP-3: Product Rat. & $P_i$[score] is the score for "$P_i$[text]" & \textcolor{blue}{concat}([\textcolor{blue}{PPEP}($P_1$), ..., \textcolor{blue}{PPEP}($P_n$)], \textcolor{gray}{", and "}). \textcolor{red}{[INPUT]} \\
            LaMP-4: News Headline & "$P_i$[title]" is the title for "$P_i$[text]" & \textcolor{blue}{concat}([\textcolor{blue}{PPEP}($P_1$), ..., \textcolor{blue}{PPEP}($P_n$)], \textcolor{gray}{", and "}). \textcolor{red}{[INPUT]} \\
            LaMP-5: Scholarly Title & "$P_i$[title]" is the title for "$P_i$[abstract]" & \textcolor{blue}{concat}([\textcolor{blue}{PPEP}($P_1$), ..., \textcolor{blue}{PPEP}($P_n$)], \textcolor{gray}{", and "})\textcolor{gray}{. Following the given patterns} \textcolor{red}{[INPUT]} \\
            LaMP-6: Email Subject & "$P_i$[title]" is the title for "$P_i$[text]" & \textcolor{blue}{concat}([\textcolor{blue}{PPEP}($P_1$), ..., \textcolor{blue}{PPEP}($P_n$)], \textcolor{gray}{", and "}). \textcolor{red}{[INPUT]} \\
            LaMP-7: Tweet Para. & "$P_i$[text]" & \textcolor{blue}{concat}([\textcolor{blue}{PPEP}($P_1$), ..., \textcolor{blue}{PPEP}($P_n$)], \textcolor{gray}{", and "}) \textcolor{gray}{are written by a person. Following the given patterns} \textcolor{red}{[INPUT]}\\\hline
    \end{tabular}
    \end{adjustbox}
    \label{tab:task-prompts}
\end{table*}

\section{Experiments}
\label{sec:exp}

\subsection{Experimental Setup}
\label{sec:exp:setup}
\subsubsection*{\textbf{Datasets}}

We adopt the LaMP benchmark \cite{lamp}--a public benchmark that encompasses a diverse set of personalized text generation tasks.\footnote{The LaMP benchmark is available at \url{https://lamp-benchmark.github.io/}.} Specifically, the benchmark comprises three personalized text classification tasks and four personalized text generation tasks. They include (1) Personalized Citation Identification (binary classification), (2) Personalized Movie Tagging (categorical classification with 15 classes), (3) Personalized Product Rating (ordinal classification from 1 to 5-star rating for e-commerce products), (4) Personalized News Headline Generation, (5) Personalized Scholarly Title Generation, (6) Personalized Email Subject Generation, and (7) Personalized Tweet Paraphrasing. 

We use the time-based separation setting offered by LaMP for data splitting. In this setting, the data for each user is split into train, development, and test sets based on their timestamp, modeling a real-world scenario in which personalized outputs for test inputs are generated using the personal documents created earlier by that user. The reason behind opting for a time-based separation in the LaMP benchmark is to investigate the impact of recency in our experiments. Table \ref{tab:task-stats} reports the statistics of the datasets.

\subsubsection*{\textbf{Evaluation Metrics}}
Following \citet{lamp}, we evaluate LaMP-1 using Accuracy, LaMP-2 using Accuracy and F1-measure, and LaMP-3 using mean absolute error (MAE) and root mean squared error (RMSE). We use ROUGE-1 and ROUGE-L \cite{lin-2004-rouge} to evaluate text generation performance on text generation datasets (LaMP-4 to LaMP-7). Statistically significant differences are identified using two-tailed paired t-test for ROUGE-1/ROUGE-L/MAE/RMSE and McNemar test for Accuracy/F1.


\begin{table*}[t]
    \centering
    \caption{The performance of our methods and the baselines on the LaMP benchmark. For all metrics, the higher values the better, except for RMSE and MAE which are used in LaMP-3. In this table, the superscript $^1$, $^2$, $^3$, $^4$, and $^5$ indicate significant improvement over No Personalization, BM25, Recency, Contriever, and RRF, respectively ($p < 0.05$).} 
    \vspace{-0.2cm}
    \resizebox{\textwidth}{!}{\begin{tabular}{l|l|c|ccc|c|cccc}
        \multirow{2}{*}{\textbf{Dataset}} & \multirow{2}{*}{\textbf{Metric}} & {\textbf{No}} & \multicolumn{4}{c|}{\textbf{Personalization Baselines}} & \multicolumn{4}{c}{\textbf{Our Methods}} \\
        & & \textbf{Personalization} & BM25 & Recency & Contriever & RRF & \ltrrl & \ltrkd & \lts-Pre & \lts-Post \\
        \hline
        
        \multirow{2}{*}{\shortstack[l]{{LaMP-1: Personalized}\\{Citation Identification}}} & \multirow{2}{*}{Accuracy $\uparrow$} & \multirow{2}{*}{0.502} & \multirow{2}{*}{0.626} & \multirow{2}{*}{0.622} & \multirow{2}{*}{0.636} & \multirow{2}{*}{0.570} & \multirow{2}{*}{{0.655$^{12345}$}} & \multirow{2}{*}{{0.668$^{12345}$}} & \multirow{2}{*}{{0.663$^{12345}$}} & \multirow{2}{*}{\textbf{0.672$^{12345}$}} \\ & & & & & & & \\
        \hline
        
        \multirow{2}{*}{\shortstack[l]{{LaMP-2: Personalized}\\{Movie Tagging}}} & Accuracy $\uparrow$ & 0.359 & 0.387 & 0.377 & 0.396 & 0.375 & 0.391$^{135}$ & 0.396$^{135}$ & 0.405$^{1235}$ & \textbf{0.430$^{12345}$}\\
        & F1 $\uparrow$ & 0.276 & 0.306 & 0.295 & 0.304 & 0.299 & 0.300$^{135}$ & 0.306$^{135}$ &  {0.314}$^{1235}$ & \textbf{0.339$^{12345}$} \\
        \hline
        
        \multirow{2}{*}{\shortstack[l]{{LaMP-3: Personalized}\\{Product Rating}}} & MAE $\downarrow$ & 0.308 & 0.298 & 0.296 & 0.299 & 0.314 & 0.286$^{145}$ & 0.290$^{15}$ & 0.282$^{12345}$ & \textbf{0.264$^{12345}$} \\
        & RMSE $\downarrow$ & 0.611 & 0.611 & 0.605 & 0.616 & 0.614 & 0.591$^{145}$ & 0.604$^{15}$ & 0.585$^{12345}$ & \textbf{0.568$^{12345}$} \\
        \hline
        
        \multirow{2}{*}{\shortstack[l]{{LaMP-4: Personalized}\\{News Headline Generation}}} & ROUGE-1 $\uparrow$ & 0.176 & 0.186 & 0.189 & 0.183 & 0.190 & 0.191$^{1}$ & 0.187$^{1}$ & 0.190$^{1}$ & \textbf{0.203$^{12345}$} \\
        & ROUGE-L $\uparrow$ & 0.160 & 0.171 & 0.173 & 0.169 & 0.176 & 0.177$^{1}$ & 0.172$^{1}$ & 0.176$^{1}$ & \textbf{0.186$^{12345}$} \\
        \hline
        
        \multirow{2}{*}{\shortstack[l]{{LaMP-5: Personalized}\\{Scholarly Title Generation}}} & ROUGE-1 $\uparrow$ & 0.478 & 0.477 & 0.475 & \textbf{0.483} & 0.478 & 0.475 & 0.477 & \textbf{0.483$^{1235}$} & {0.480} \\
        & ROUGE-L $\uparrow$ & 0.428 & 0.427 & 0.426 & \textbf{0.433} & 0.428 & 0.427 & 0.428 & 0.431$^{1235}$ & {0.429} \\
        \hline
        
        \multirow{2}{*}{\shortstack[l]{{LaMP-6: Personalized}\\{Email Subject Generation}}} & ROUGE-1 $\uparrow$ & 0.335 & 0.412 & 0.403 & 0.401 & 0.394 & 0.394 & 0.415$^{1345}$ & 0.426$^{12345}$ & \textbf{0.433$^{12345}$} \\
        & ROUGE-L $\uparrow$ & 0.319 & 0.398 & 0.389 & 0.386 & 0.381 & 0.381 & 0.400$^{1345}$ & 0.411$^{12345}$ & \textbf{0.418$^{12345}$} \\
        \hline
        
        \multirow{2}{*}{\shortstack[l]{{LaMP-7: Personalized}\\{Tweet Paraphrasing}}} & ROUGE-1 $\uparrow$ & {0.449} & 0.446 & 0.444 & 0.440 & 0.446 & 0.448$^{4}$ & 0.441 & 0.450$^{2345}$ & \textbf{0.461$^{12345}$} \\
        & ROUGE-L $\uparrow$ & 0.396 & 0.394 & 0.393 & 0.390 & 0.395 & 0.397$^{4}$ & 0.391 & 0.400$^{2345}$ & \textbf{0.409$^{12345}$} \\
        \hline
    \end{tabular}}
    \label{tab:main-results}
\end{table*}

\subsubsection*{\textbf{Training Configurations}}

An integral part of our training pipeline is the evaluation function $\textsc{Eval}$. We use the standard metrics suggested by the LaMP benchmark for each dataset to implemented the $\textsc{Eval}$ function. In more detail, we measure accuracy for the binary and categorical classification tasks (LaMP-1 and LaMP-2) and ROUGE-1 \cite{lin-2004-rouge} for the text generation tasks (LaMP-4, LaMP-5, LaMP-6, and LaMP-7). Given that the evaluation metric for LaMP-3 is MAE (Mean Absolute Error), where lower values are preferable, we need to adapt the evaluation function to use it as a reward function. The modification is as follows: 
\begin{equation}
    \textsc{Eval}_{\text{LaMP-3}}(y, \hat{y}) = \frac{\max(|1-y|, |5-y|) - \text{MAE}(y, \hat{y})}{\max(|1-y|, |5-y|)}
\end{equation}
where $\hat{y}$ is the prediction and $y$ is the target output. This reward function normalizes the MAE score by measuring its distance from the worst score achievable based on the model's prediction.\footnote{According to the implementation of the LaMP benchmark, the worst score attainable by a model in the LaMP-3 task is denoted as $\max(|1-y|, |5-y|)$.} In this scenario, a correct prediction by the model results in a score of 1, while in the worst-case prediction, it receives a score of 0. 

In this paper, we use the Adam optimizer \cite{adam} with a learning rate of $10^{-5}$. We dedicate 5\% of the training steps to warmup with a linear scheduler. We also use gradient clipping with the value of 1. To accommodate the task requirements, we set the maximum input and output lengths to 512 tokens for LLMs following \citet{lamp}. However, we use the maximum input length of 1024 for retrieval selection in order to incorporate a prompt and the corresponding output from the LLM. We train the retrieval models for 10 and the retriever selection models for 20 epochs. In all experiments, following \citet{lamp}, we utilize FlanT5-XXL\cite{flant5}-- an instruction-tuned open-source LLM with 11B parameters. We use a beam size of 4 in beam search for text generation \cite{beam-search}. The effective batch size in all experiments is set to 64 (8 accumulation steps with batch size 8). We performed all the experiments on a single A100 Nvidia GPU with 80GB memory and 128GB of RAM. 

In all experiments, following \citet{lamp}, we use the non-template parts of the LLM input $x$ as the query for personal document retrieval. We followed \citet{lamp} for prompt templates (i.e., $\phi_p$) for each dataset in LaMP. They are listed in \tablename~\ref{tab:task-prompts}. In all experiments, the process of creating personalized prompts for evaluating models involves retrieving four items from the user profile. In crafting documents from each user profile, we adhere to the approach established in \cite{lamp}, appending the date of the document to it. The date is prefixed with \textit{date: [date].} For implementing BM25, we use the \texttt{rank\_bm25} library.\footnote{\url{https://github.com/dorianbrown/rank_bm25}} All the neural models in this paper are implemented using the PyTorch library \cite{pytorch}.

\subsubsection*{\textbf{Baseline Methods}}
We compare the proposed approaches against the following retrieval models for personalized text generation. 
\begin{itemize}[leftmargin=*]
    \item \textbf{No Personalization:} We employ FlanT5-XXL \citep{flant5} as a non-personalized baseline. In this baseline, the model is presented with the original task's input without any modification.
    \item \textbf{Personalized Baselines:} Following \citet{lamp}, we utilize BM25 \citep{bm25}, Recency, and Contriever \citep{contriever} to retrieve items from the user profile for LLM personalization. There are, of course, many other neural ranking models that may outperform these baselines on some retrieval benchmarks. However, it is important to note that our optimization approaches can be applied to any neural ranking model, including any missing baseline from this list. That being said, we do not aim at comparing different model architectures, instead we aim at demonstrating the impact of our optimization methods. Note that no other retrieval models have ever been used on LaMP and, to the best of our knowledge, this list consists of all methods in the retrieval-augmented LLM personalization literature.
    Furthermore, we apply Reciprocal Rank Fusion (RRF) \citep{rrf} to integrate the retrieval lists generated by all the retrievers. This fusion-based approach is employed for comparison with our retriever selection method. Subsequently, we employ these retrieved items to formulate a personalized input prompt for FlanT5-XXL. 
    
\end{itemize}

\subsection{Empirical Results}
This section provides empirical evidence to answer research questions that shed light into the proposed approaches in this paper.

\noindent \subsubsection*{\textbf{How does personalization using the proposed approaches impact text generation performance?}}
To answer this question, we compare our methods with the non-personalized baseline, i.e., FlanT5-XXL without augmentation with personal information. Table \ref{tab:main-results} presents the results on the LaMP benchmark. LLM Personalization using both \ltrrl and \ltrkd improves the performance on LaMP-1, LaMP-2, LaMP-3, LaMP-4, and LaMP-6.  After applying the retrieval model selection methods (\lts-Pre and \lts-Post), the non-personalized model is beaten on all datasets and in terms of all metrics. The performance gains are statistically significant in almost all cases. This is an important finding in the sense that no personalized baseline could perform better than a non-personalized LLM on LaMP-7, while \lts-Pre and \lts-Post demonstrate that personalized LLMs can ultimately demonstrate performance gain on these datasets. This finding also suggests the impact of retrieval augmentation for the purpose of LLM personalization.

\noindent \subsubsection*{\textbf{How do \ltr optimization algorithms impact text generation performance?}}
To answer this, we must compare \ltrrl and \ltrkd with the Contriever baseline, since they are initialized with the Contriever model and fine-tuned using our proposed \ltrrl and \ltrkd algorithms. The results in \tablename~\ref{tab:main-results} suggest that applying \ltrrl to Contriever yields performance gain on LaMP-1, LaMP-3, LaMP-4, and LaMP-7. \ltrkd additionally outperforms Contriever on LaMP-6.  Notably \ltrkd performs better than \ltrrl on the tasks with binary feedback from the language model (LaMP-1 and LaMP-2). Comparing \ltrrl and \ltrkd suggests there is no clear winner among them. This once again attests that each personalization task have different requirements, thus motivating the need for retrieval selection in retrieval-augmented LLM personalization.

\begin{table}[t]
    \centering
    \caption{The success rate of models in selecting the best performing retriever for each input. The superscript $^*$ indicates significant improvement over the best baseline ($p<0.05$). }
    \vspace{-0.4cm}
    \resizebox{\linewidth}{!}{
    \begin{tabular}{l|cccc|cc}
        \multirow{2}{*}{\textbf{Dataset}} & \multicolumn{6}{c}{\textbf{Success Rate}} \\
        & WIG & NQC & $\sigma_{\text{max}}$ & $\sigma_{\text{50\%}}$ & \lts-Pre & \lts-Post \\
        \hline
        \multirow{2}{*}{\shortstack[l]{{LaMP-1: Personalized}\\{Citation Identification}}} & \multirow{2}{*}{0.858} & \multirow{2}{*}{0.824} & \multirow{2}{*}{0.848} & \multirow{2}{*}{0.847} & \multirow{2}{*}{0.865$^*$} & \multirow{2}{*}{\textbf{0.874$^*$}}\\ & & & & & & \\
        \hline 
        \multirow{2}{*}{\shortstack[l]{{LaMP-2: Personalized}\\{Movie Tagging}}} & \multirow{2}{*}{0.908} & \multirow{2}{*}{0.918} & \multirow{2}{*}{0.910} & \multirow{2}{*}{0.910} & \multirow{2}{*}{0.936$^*$}  & \multirow{2}{*}{\textbf{0.962$^*$}} \\ & & & & & & \\
        \hline
        \multirow{2}{*}{\shortstack[l]{{LaMP-3: Personalized}\\{Product Rating}}} & \multirow{2}{*}{0.896} & \multirow{2}{*}{0.890} & \multirow{2}{*}{0.896} & \multirow{2}{*}{0.894} & \multirow{2}{*}{0.903} & \multirow{2}{*}{\textbf{0.920$^*$}} \\& & & & & & \\
        \hline
        \multirow{2}{*}{\shortstack[l]{{LaMP-4: Personalized}\\{News Headline Generation}}} & \multirow{2}{*}{0.401} & \multirow{2}{*}{0.404} & \multirow{2}{*}{0.394} & \multirow{2}{*}{0.398} & \multirow{2}{*}{0.401} & \multirow{2}{*}{\textbf{0.447$^*$}} \\ & & & & & & \\
        \hline
        \multirow{2}{*}{\shortstack[l]{{LaMP-5: Personalized}\\{Scholarly Title Generation}}} & \multirow{2}{*}{0.562} & \multirow{2}{*}{0.572} & \multirow{2}{*}{0.562} & \multirow{2}{*}{0.557} & \multirow{2}{*}{\textbf{0.600$^*$}} & \multirow{2}{*}{0.577} \\ & & & & & & \\
        \hline
        \multirow{2}{*}{\shortstack[l]{{LaMP-6: Personalized}\\{Email Subject Generation}}} & \multirow{2}{*}{0.613} & \multirow{2}{*}{0.606} & \multirow{2}{*}{0.614} & \multirow{2}{*}{0.617} & \multirow{2}{*}{0.633$^*$} & \multirow{2}{*}{\textbf{0.641$^*$}} \\ & & & & & & \\
        \hline
        \multirow{2}{*}{\shortstack[l]{{LaMP-7: Personalized}\\{Tweet Paraphrasing}}} & \multirow{2}{*}{0.851} & \multirow{2}{*}{0.840} & \multirow{2}{*}{0.852} & \multirow{2}{*}{0.851} & \multirow{2}{*}{0.860}  & \multirow{2}{*}{\textbf{0.898$^*$}} \\ & & & & & & \\
        \hline
    \end{tabular}
    }
    \label{tab:retrieval-model-selection-results}
    \vspace{-0.6cm}
\end{table}

\begin{table*}[t]
    \centering
    \caption{Studying the impact of \ltr algorithms on the end-to-end performance of our pipeline with retrieval model selection. In this table, the superscript $^*$ indicates significant improvement over w/o \ltr approach ($p<0.05$).}
    \vspace{-0.3cm}
    \resizebox{\textwidth}{!}{\begin{tabular}{l|l|cc|cc|cc}
        \multirow{2}{*}{\textbf{Dataset}} & \multirow{2}{*}{\textbf{Metric}} & \multicolumn{2}{c|}{\textbf{\lts-Pre}} & \multicolumn{2}{c|}{\textbf{\lts-Post}} & \multicolumn{2}{c}{\textbf{Oracle}} \\
        & & w/o \ltr & w/ \ltr & w/o \ltr & w/ \ltr & Lower-bound & Upper-bound \\
        \hline
        {\shortstack[l]{{LaMP-1: Personalized Citation Identification}}} & {Accuracy $\uparrow$} & {0.646} & {0.663$^*$} & {0.644} & {0.672$^*$} & {0.381} & {0.798} \\
        \hline
        
        \multirow{2}{*}{\shortstack[l]{{LaMP-2: Personalized Movie Tagging}}} & {Accuracy $\uparrow$ } & 0.403 & 0.405 & {0.425} & 0.430 & {0.296} & {0.468} \\
        & {F1  $\uparrow$} & 0.311 & 0.314 & 0.330 & 0.339 & {0.216} & {0.380} \\
        \hline

        \multirow{2}{*}{\shortstack[l]{{LaMP-3: Personalized Product Rating}}} & {MAE $\downarrow$ } & 0.287 & 0.282 & 0.269 & 0.264 & {0.450} & {0.181} \\
        & {RMSE  $\downarrow$}  & 0.595 & 0.585  & 0.577 & 0.568 & {0.772} & {0.462} \\
        \hline

        \multirow{2}{*}{\shortstack[l]{{LaMP-4: Personalized News Headline Generation}}} & {ROUGE-1 $\uparrow$ } &  0.189 & 0.190 & {0.191} & 0.203$^*$ & {0.103} & {0.269} \\
        & {ROUGE-L  $\uparrow$} & 0.174 & 0.176 & 0.177 & 0.186$^*$ & {0.098} & {0.243} \\
        \hline
        
        \multirow{2}{*}{\shortstack[l]{{LaMP-5: Personalized Scholarly Title Generation}}} & {ROUGE-1 $\uparrow$ } & 0.478 & 0.483$^*$ & {0.475} & 0.480 & {0.388} & {0.548} \\
        & {ROUGE-L  $\uparrow$} & 0.428 & 0.431$^*$ & 0.427 & 0.429 & {0.349} & {0.492} \\
        \hline

        \multirow{2}{*}{\shortstack[l]{{LaMP-6: Personalized Email Subject Generation}}} & {ROUGE-1 $\uparrow$ } &  0.426 & 0.426 & {0.394} & 0.433$^*$ & {0.239} & {0.511} \\
        & {ROUGE-L  $\uparrow$} &  0.412 & 0.411 & 0.381 & 0.418$^*$ & {0.228} & {0.492} \\
        \hline

        \multirow{2}{*}{\shortstack[l]{{LaMP-7: Personalized Tweet Paraphrasing}}} & {ROUGE-1 $\uparrow$ } & 0.449 & 0.450 & {0.448} & 0.461$^*$ & {0.410} & {0.470} \\
        & {ROUGE-L  $\uparrow$} & 0.398 & 0.400 & 0.397 & 0.409$^*$ & {0.361} & {0.417} \\
        
        \hline
    \end{tabular}}
    \label{tab:ablation-components}
\end{table*}

\noindent \subsubsection*{\textbf{How effective are the proposed retrieval selection methods?}}
To assess the efficacy of retriever selection, we measure its success rate in selecting the best performing retriever in the retriever pool $\mathbf{R}$ for each input. Note that for inputs with multiple best-performing retrieval models, a selection is considered successful if any of them is selected. In this experiment, we incorporated several unsupervised Query Performance Prediction (QPP) methods, including WIG~\cite{wig}, NQC~\cite{nqc}, and $\sigma_{\text{max}}$ and $\sigma_{\text{x\%}}$ \cite{sigma-qpp}, to perform a comparative analysis with our proposed method. To accomplish this, we assign the task to each QPP approach of providing a score to the retrieved results for each retriever. The retriever with the highest score is then selected for that particular input. It is important to note that since Recency does not provide score for retrieved results, we consider the reciprocal rank of each document as its score. In no personalization retriever, we assign a score of zero to all items in the profile. It is crucial to highlight that the utilization of supervised QPP methods, such as BERT-QPP \cite{bert-qpp}, was not feasible due to their reliance on query-document relevance labels, which are unavailable in our datasets.

The results in Table~\ref{tab:retrieval-model-selection-results} demonstrate the superior performance of both the \lts-Pre and \lts-Post across nearly all datasets. Specifically, \lts-Post achieves a significant improvement over all baselines for the all datasets, except for the LaMP-5 dataset. In this dataset, \lts-Pre shows a significant improvement compared to the baselines. Likewise, \lts-Pre outperforms all baselines in all datasets except LaMP-4, with significant improvements observed in LaMP-1, LaMP-2, LaMP-5, and LaMP-6. Overall, the outcomes of this experiment suggest that the proposed approach for retrieval model selection consistently outperforms the baselines.

The results also indicate that for all classification datasets (i.e., LaMP-1, LaMP-2, and LaMP-3) and LaMP-7 for generation, both pre- and post-generation models achieve a success rate of over 80\%. This suggests that the model has to some extent successfully learned to choose the most suitable retriever for each input. Conversely, for the remaining text generation datasets (i.e., LaMP-4, LaMP-5, and LaMP-6), the accuracy is lower, ranging from 40\% to 65\%. We attribute this to the inherent complexity of text generation tasks compared to text classification. Comparing \lts-Pre and \lts-Post, the latter exhibits higher success rate in all tasks except LaMP-5. This suggests that employing the generated output in retrieval selection can have a substantial impact on the performance. 

Looking back to the results in \tablename~\ref{tab:main-results}, the post-generation retrieval selection model (\lts-Post) performs better than the pre-generation selection model (\lts-Pre) on six out of seven datasets; LaMP-5 is the exception, which is explained by the results in \tablename~\ref{tab:retrieval-model-selection-results}.
\lts-Pre consistently outperforms the baselines significantly in all classification tasks (LaMP-1, LaMP-2, and LaMP-3) as well as in LaMP-6. In the remaining datasets, \lts-Pre achieves comparable or superior results to the baselines, although the differences are not statistically significant. Finally, \lts-Post leads to the best personalized text generation performance on six out of seven datasets. 

\noindent \subsubsection*{\textbf{What is the method that results in the highest personalized text generation performance?}}
According to \tablename~\ref{tab:main-results}, \lts-Post performs best on six out of seven datasets (all but LaMP-5). Contriever, on the other hand, demonstrates the highest performance on LaMP-5. The performance gains by \lts-Post on all the remaining six datasets are statistically significant, according to a two-tailed paired t-test for generation and ordinal classification datasets and McNemar test for binary and categorical text classification datasets.

\noindent \subsubsection*{\textbf{What is the impact of \ltr algorithms on retrieval selection results?}}
To investigate the impact of training personalized retrievers using the proposed \ltr algorithms (i.e., both \ltrrl and \ltrkd) on the end-to-end performance of the pipeline with retrieval selection, we conduct an analysis by excluding the fine-tuned retrievers from the retriever pool $\mathbf{R}$. The outcomes of this experiment along with the Oracle performance (both lower and upper bound for retrieval selection) are reported in Table~\ref{tab:ablation-components}. The results indicate that the models without \ltr in its retriever pools achieve lower performance on all datasets in both pre- and post-generation settings. The only exception is the \lts-Pre results on LaMP-6, where including \ltr algorithms does not make any significant impact. This suggests that our ultimate performance gain is not just because of the retrieval selection models; Instead, the retrieval optimization approaches presented in Section \ref{sec:personalized-retriever} are effective in enhancing the end-to-end performance of the pipeline.

Finally, a comparison between our results and the Oracle performance in Table \ref{tab:ablation-components} provides insight into the potential for further improvements in retrieval selection. For instance, in LaMP-3 and LaMP-4, our best performing method only achieves 68.3\% and 75.4\% of the Oracle's upper-bound, respectively. This suggests that there are still substantial room for improvement. However, the corresponding numbers for LaMP-1, LaMP-2, LaMP-5, LaMP-6, and LaMP-7 are 84.2\%, 91.8\%, 88.1\%, 84.7\%, and 98.0\%, respectively, suggesting that our best performing retrieval selection method is performing very close to the upper-bound performance.

\section{Conclusions and Future Work}

This paper explored personalization of LLMs through a retrieval augmentation pipeline with a focus on optimizing the retrieval component. We introduced two solutions for optimizing ranking models by soliciting personalized feedback from the language model, one based on reinforcement learning where the reward function is defined based on the personalized text generation quality, and another based on knowledge distillation from the language model to the retrieval model. Subsequently, we observed that personalization tasks can benefit from different retrieval models, depending on specific needs and requirements. Given this observation, we developed a pre-generation and a post-generation retriever selection model. Evaluation on seven diverse personalization tasks from the LaMP benchmark showed that our proposed methods outperform competitive baselines on six out of seven datasets with statistically significant improvements. Through careful ablation studies, we demonstrate the impact of each component used our pipeline. 

In this work, we solely focused on optimizing and selecting the ranking models for LLM personalization. One limitation of this work lies in the use of static templates for generating prompts for the LLM. In the future, we aim at optimizing the prompt generation component using the feedback obtained from the downstream LLM performance. In addition, all datasets in the LaMP benchmark focus on short text generation tasks. We will explore personalized long text generation methods in the future.

\section*{Acknowledgment}

This work was supported in part by the Center for Intelligent Information Retrieval, in part by Lowe’s, in part by an Amazon Research Award, Fall 2022 CFP, in part by an award from Google, and in part by and award from Microsoft. Any opinions, findings and conclusions or recommendations expressed in this material are those of the authors and do not necessarily reflect those of the sponsor.

\balance
\bibliographystyle{ACM-Reference-Format}
\bibliography{XX-references}

\end{document}